# Investigating a Hybrid Metaheuristic for Job Shop Rescheduling

Salwani Abdullah[2], Uwe Aickelin[1], Edmund Burke[1], Aniza Mohamed Din[1] and Rong Qu[1]

[1] School of Computer Science
University of Nottingham
Nottingham, UK
{uxa, ekb, amd, rxq}@cs.nott.ac.uk

[2] Faculty of Information Science and Technology
Universiti Kebangsaan Malaysia
Bangi, Selangor, Malaysia
salwani@ftsm.ukm.my

**Abstract.** Previous research has shown that artificial immune systems can be used to produce robust schedules in a manufacturing environment. The main goal is to develop building blocks (antibodies) of partial schedules that can be used to construct backup solutions (antigens) when disturbances occur during production. The building blocks are created based upon underpinning ideas from artificial immune systems and evolved using a genetic algorithm (Phase I). Each partial schedule (antibody) is assigned a fitness value and the best partial schedules are selected to be converted into complete schedules (antigens). We further investigate whether simulated annealing and the great deluge algorithm can improve the results when hybridised with our artificial immune system (Phase II). We use ten fixed solutions as our target and measure how well we cover these specific scenarios.

## 1 Introduction

Job shop scheduling problems are concerned with tackling the problem of assigning $n$ jobs to $m$ machines and are very well studied. The problem has been addressed using several local search techniques such as tabu search, genetic algorithms and simulated annealing as observed by Jain and Meeran in [20] who analysed some of the techniques used and made comparisons between them. In this paper, we are specifically trying to tackle the problem of changing job shop environments. Such changes include the unexpected arrival dates of jobs into the factory. If jobs arrive too early, it could lead to them being stored for long periods of time and if they arrive late, it could cause delays in processing other jobs [12,21]. An efficient method of rescheduling is needed to manage the problem.

## 2 Problem Description

Job shop schedules require constant revision as problems could happen during production and delays could cost money and time. For a detailed discussion of job shop problems, see [4] and [24]. Rescheduling is important to ensure the production can maintain its flow and minimize interruption. A quick solution to such a problem is usually very much preferred compared to starting from scratch. This observation represents the motivation for this paper. It addresses the goal of being able to generate a diverse range of partial schedules that could be used as a replacement in the event of changes in a job shop environment. These partial schedules should enable us to generate a new complete schedule in order to keep the manufacturing process flowing smoothly with a low level of interruption. In this paper, we will employ an artificial immune system algorithm to build these partial schedules. We use previous, complete schedules (later known as the antigen universe) to build a collection of partial schedules. This data stems from [18]: the number of jobs used is 15, assigned to five machines. We employ precedence constraints to the jobs when building the partial schedules. These partial schedules are then evolved using a genetic algorithm. These processes will be explained in Section 3.1. In Section 3.2, we hybridise the newly developed artificial immune system with local search to see whether there is improvement to the results.

## 3 A Hybrid Metaheuristic Model

Artificial immune systems (AIS) are motivated by immunology. The biological immune system defends the body from antigens. It generates antibodies that can attack specific antigen. An overview of artificial immune systems research can be seen in [2] and [7].

Previous research on AIS for scheduling has shown that an AIS model can be used in a job shop setting. Different scheduling problems have been addressed including the job shop scheduling problem [5,6,13], the hybrid flow shop scheduling problem [11] and the job shop rescheduling problem [16,17,18], which is the main concern of this paper. Hart and Ross [17], in their research, tackled this problem by building a block of partial schedules. There are many definitions given to the antibody and the antigen for the problem, which are used to build the partial schedules. We are employing the definition given by Hart and Ross in [17]. The key definitions used in this research are outlined below:

- An **antigen** is defined as "*the sequence of jobs on a particular machine given a particular scenario*" [17], which represents a full schedule for the problem. The antigens are represented by a sequence of numbers of length 15 for the problem tested here.
- An **antibody** is defined as "*a short sequence of jobs that is common to more than one schedule*" [17], which is also known as a partial schedule. The antibodies are represented by sequences of numbers of length 5, where the length of an antibody is less than the length of an antigen.

- An **antigen universe** is considered to be a collection of antigens (sequence of jobs) to be matched with the antibodies (partial schedules). An antigen universe has to be prepared before we can build an antibody population.
- An **antibody population** is a collection of partial schedules constructed from gene libraries.
- **Gene libraries** consist of genotypes [19,23]. The gene libraries in this research are constructed from all the antigens in the antigen universe.
- A **final population** consists of a collection of best antibodies. When we hybridise our AIS model with a local search method, the final population from our AIS model will be the initial solutions to the local search method.
- **Fitness** represents the value assigned to each antibody in the antibody population to evaluate the coverage of an antibody over the antigens. The higher the fitness, the better an antibody will be.

We have divided our work into two phases. In the first phase, an AIS model is used to generate the antibody population, with $l = 5$ where $l$ is the length of an antibody. A genetic algorithm is then used to evolve the antibodies. The idea is that only the antibodies with the highest fitness (i.e. the best antibodies) that have most of the jobs matched with the antigens will be kept in the final population. It is important to note here that we used a genetic algorithm to evolve the antibody population as used in [17,18]. We modified the algorithm in [17,18] with the aim of improving the results. The simulated annealing and great deluge algorithms are then applied respectively in the second phase by using the best antibodies selected in the first phase (final population) as initial solutions. In this research, we are using the parameters adapted from [3]. The aim is to investigate if we can improve the fitness of the antibodies developed in the final population in Phase I as both local search methods have been known to produce good results for other scheduling problems such as examination timetabling (e.g. [3,8]).

### 3.1 Phase I: The Artificial Immune System Model

Before we generate the antibody populations, we need to have an antigen universe. The antigen universe for this research is the same as that used by Hart and Ross [18], which is based on a benchmark problem by Morton and Pentico [22]. The number of jobs used in this problem is 15 and the jobs have to be assigned to five machines. Hart and Ross created ten test scenarios by mutating the arrival dates of the jobs to a random date between 0 – 300 with a probability of 0.2. The arrival dates must not be less than $p_t$ days before the due date, where $p_t$ is the processing time of the job. A genetic algorithm developed by Fang et al [12] is used to generate five schedules for each of these test-scenarios. This resulted in five sets of ten schedules; one for each machine, and these schedules became the antigen universe for this research. This research uses the antigen universe generated from one of the machines with the assumption that all machines have a similar pattern of jobs.

**Generating the Antibody Population.** The first step in this model is to generate an antibody population (a collection of antibodies) from gene libraries [6,17,18,26]. The gene libraries in this research are constructed from all the antigens in the antigen universe. The antigens are divided into five libraries, where each library consists of ten partial schedules of size 3 also known as components. An antibody for this research is constructed based on a modular design method [14,19,23,25] where the length of each antibody is 1/3 the length of each antigen. We are using a small size problem because we are interested in evolving the antibodies using the AIS algorithm and hybridising the model with a local search method to see if we can improve the results. In our future work, a larger size of problem will be used.

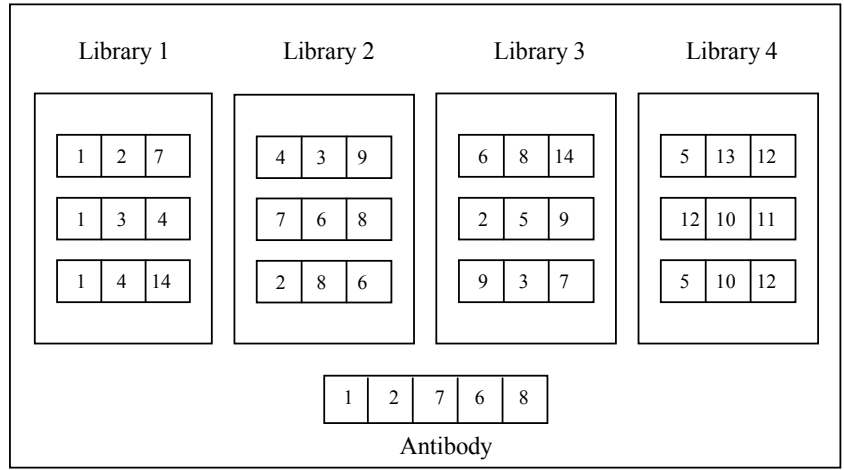

**Fig. 1.** Constructing an antibody from gene libraries

In the example in Figure 1, the gene libraries consist of four libraries and each library contains three components. Three jobs are allocated in each component. Following the modular design method, there are several ways to combine the components from gene libraries to produce an antibody. In Figure 1, we select the first component from Library 1 and combine it with the second component from Library 2 to produce an antibody. Since the length of an antibody is 5 jobs, a possible combination between components in Library 1 and Library 2

$$C\binom{n_1+n_2}{r_1+r_2} = \frac{(n_1+n_2)!}{(n_1+n_2-r_1-r_2)!(r_1+r_2)!}. \quad (1)$$

can be constructed from this example, where $n_1$ and $n_2$ represent the number of jobs in the components from the first and second library, respectively, and $r_1$ and $r_2$ represent the number of jobs to be selected from the components ($r_1 + r_2 = 5$). In Figure 1, we can see a combination of three jobs from the first component and two jobs from the second component. Therefore, jobs 1, 2 and 7 from the first component in Library 1 are combined with jobs 6 and 8 selected from the second component in Library 2. We can get other combinations from these two components using (1) above to generate an antibody population. This process is repeated until all the components in Library 1 have been combined with all the components in Library 2 as well as all the other libraries.

We also have to ensure that there will be no duplicate jobs in the antibody. We compare each antibody generated in the population and eliminate the antibodies with duplicated jobs. The process will go on until a population of antibodies is generated. By doing this, we develop a level of antibody diversity.

We generate three types of antibody populations:
1. A population with antibody duplication (we can have several similar antibodies in one population) – Type A (4514 antibodies)
2. A population with no antibody duplication regardless of the source gene libraries (no similar antibodies in one population) – Type B (2416 antibodies)
3. A population with antibody duplication (only when the antibodies are constructed from different source libraries) – Type C (2839 antibodies)

We generated these versions to see whether having a large number of similar antibodies in one population would affect the coverage of the antigen universe by the antibody population. An initial antibody population of size 100 is selected randomly.

**The Matching Function.** A matching function is used as the evaluation function within the genetic algorithm to calculate the fitness of each antibody in the antibody population. A sample of antigens is first selected from the antigen universe. Each antibody is then matched against each of the antigens selected by aligning an antigen string with an antibody string and calculating a matching score.

| Antigen | 1 | 2 | 7 | 4 | 3 | 9 | 6 | 8 | 14 | 5 | 13 | 12 | 10 | 11 | 15 | Match Score |
|---|---|---|---|---|---|---|---|---|---|---|---|---|---|---|---|---|
| Antibody | 4 | 3 | 9 | 5 | 12 | | | | | | | | | | | 0 |
| | | 4 | 3 | 9 | 5 | 12 | | | | | | | | | | 0 |
| | | | 4 | 3 | 9 | 5 | 12 | | | | | | | | | 0 |
| | | | | **4** | **3** | **9** | 5 | 12 | | | | | | | | 15 |
| | | | | | 4 | 3 | 9 | 5 | 12 | | | | | | | 0 |
| | | | | | | 4 | 3 | 9 | 5 | 12 | | | | | | 0 |
| | | | | | | | 4 | 3 | 9 | **5** | 12 | | | | | 5 |
| | | | | | | | | 4 | 3 | 9 | 5 | **12** | | | | 5 |
| | | | | | | | | | 4 | 3 | 9 | 5 | 12 | | | 0 |
| | | | | | | | | | | 4 | 3 | 9 | 5 | 12 | | 0 |
| | | | | | | | | | | | 4 | 3 | 9 | 5 | 12 | 0 |

**Fig. 2.** The process of matching an antibody with an antigen by aligning the antibody at every possible alignment position

Based on the example in Figure 2, if there is an antigen string '1 2 7 4 3 9 6 8 14 5 13 12 10 11 15', and an antibody string '4 3 9 5 12', we have to align the antibody at every possible alignment position with the antigen gene by gene in order to calculate a matching score. A matching score is calculated by summing up the scores from the matches where a match of each position contributes a score of five. Therefore, based on the number of matches between both the antibody and the antigen, the matching score for the example given above is 15, which is the best possible match found (highest matching score) by this process. Since an antibody is matched with each of the antigens in the sample, if the antibody is matched against more than one antigen, a total matching score for the antibody is arrived at by summing up the highest matching scores of matching the antibody with each of the antigens in the previous process.

Hart and Ross [17] selected certain samples of antibodies from the antibody population to be matched with a sample of antigens and repeated the matching process for a certain number of iterations based on the number of antigens selected. In our algorithm, we matched all the antibodies in the population with the antigens and ran the matching process only once. We would also like to note that, for our preliminary experiments, we did not include any wildcard genes in any antibody in the antibody population as we wanted to see the exact fitness of the antibodies as we matched them with the antigens. In [17], the authors allow a wild card match between the antibody and the antigen. A wild card is used as a substitute to any job.

**Crossover and Mutation.** A genetic algorithm was implemented based on GENESIS [15] and this was used to evolve the antibody population. We used an order-based crossover operator, as it can ensure no job duplication in an antibody for any relationship between two parent antibodies. During crossover, we applied tournament selection to select the best antibody to be included in the next generation. We evaluated the fitness of the children produced and compared their fitness with the fitness of the parents. If the children had lower fitness than the parents, they were discarded, and the parents were selected for inclusion in the next generation. Only the best antibodies, i.e. antibodies with the highest fitness, were considered for the next generation. A mutation operator, which randomly mutates each gene with a probability of 0.2, was also applied as used in [17].

### 3.2 Phase II: Simulated Annealing and the Great Deluge Algorithm

In the second phase, we apply local search methods on the final population generated from the first phase to improve the fitness of the antibodies.

**Simulated Annealing.** The simulated annealing algorithm is well studied and an overview and description is presented in [1].

As mentioned above, the initial solution for this algorithm is provided by the final population developed using the model described in Section 3.1. We set the initial temperature $T_0$ to 5000 and the final temperature $T_f$ to 0.05. The temperature will be decreased by $\alpha$, where $\alpha$ is defined as 0.98 which has been found to be an effective value in the literature [8,9,27].

While the current temperature is greater than the final temperature, new antibodies, $Ab_{new}$ are generated. This is done by applying two different operators, respectively in two different experiments; changing one job in $Ab$ or swapping two jobs in $Ab$, where $Ab$ represents the antibodies in the antibody population. The fitness of each antibody is then calculated using the same matching function as applied in the artificial immune system model. The new antibody will be kept if the fitness of the new antibody is better than the fitness of the current best antibody in the antibody population. Otherwise, it is accepted with a probability of $e^{-\delta/T}$. Here, $\delta$ is defined as the difference between the fitness of the new antibody and the old antibody. We also record the best antibody found overall. This is included in the antibody population (final solution) if it is better than the original antibody.

**The Great Deluge Algorithm.** Dueck [10] introduced the Great Deluge algorithm in 1993. This algorithm is similar to Simulated Annealing but it has a different acceptance process for worse solutions. The control parameter in this algorithm is called a level or boundary. A worse solution is still acceptable as long as it is within the boundary, which, at the beginning, is set to the fitness of the initial solution. The boundary is then decreased by a fixed decay rate, $\beta$, at every iteration of the search.

The initial solution for this algorithm is also provided by the final population generated using the artificial immune system model in Phase I. Here we set the number of iterations, *iter* to 120, which is the possible number of new antibodies generated by an antibody. We also set the estimated quality of the final solution, *f(EQ)*, which is the maximum fitness value for an antibody, depending on the number of antigens selected in the matching function. If the number of antigens selected is one, the maximum fitness value for *f(EQ)* is 25. This estimated quality represents the final estimated fitness value of an antibody. The boundary to the fitness of each antibody known as *boundary* is decreased by a decreasing rate, $\beta$ [3] which is defined as follows:

$$\beta = (f(Ab) - f(EQ)) / iter .  \qquad (2)$$

While the number of iterations does not exceed *iter*, new antibodies are generated by using the same two operators used in the simulated annealing algorithm. We then calculate the fitness of each new antibody generated, *f(Ab)*, by using the same matching function as described in Phase I. A new antibody which is worse than the old one will only be accepted if its fitness is less than the boundary. This loop will also stop if there is no more improvement within a fixed number of iterations.

## 4  Experiments and Results

As described in Section 3.1, Hart and Ross created ten test scenarios from a base problem, *jb11,* taken from Morton and Pentico [18,22] and the schedules generated from the problem became the antigen universe for this research. We generate three types of antibody populations in order to determine whether having a large number of similar antibodies in one population would affect the coverage of the antigen universe by the antibody population. Our program was coded in C and the experiments were executed on a PC in Windows XP environment with a Pentium 4-2.4 GHz processor and 512 MB RAM.

In the first phase, an initial population of size 100 was selected randomly for each type of antibody population and these populations were evolved using a genetic algorithm for 250 generations, with a crossover rate of 0.7 as used in [17]. We used two mutation rates in the experiments. A mutation rate of 0.2 is employed as it is the same parameter used in [17] and, therefore, it is easier for us to make a comparison with those results. We then used a mutation rate of 0.001 as this gave us a steady growth of the fitness of the antibodies in the antibody population. The antibodies evolved here were the antibodies with the highest fitness value in each generation. At the end of the generation, the antibody library should consist of a collection of general and specific antibodies, which could either match many antigens or only one specific antigen.

Tables 1 and 2 show the average number of antigens that cannot be matched by any antibody for a matching threshold ranging from 2 to 5. A matching threshold, $t_m$, is a guideline on when we can determine whether an antibody and antigen are matched. The number of genes to bind or match must be greater or equal to the threshold value of $t_m$ [17]. This experiment tests the coverage of the antigen universe by the antibody population. Table 1 shows the results of the experiment by Hart and Ross [17]. Table 2 shows the results of our experiments performed on final populations generated from the antibody population Type A, Type B and Type C, respectively (from Phase I) with a mutation rate of 0.2.

**Table 1.** Average number of antigens (out of a possible 10) not matched by any antibody as generated by Hart and Ross [17]

| Match Threshold | Ag = 1 | | | Ag = 4 | | | Ag = 8 | | |
| --- | --- | --- | --- | --- | --- | --- | --- | --- | --- |
| | Ab | | | Ab | | | Ab | | |
| | 5 | 10 | 30 | 5 | 10 | 30 | 5 | 10 | 30 |
| 2 | 0.9 | 0.0 | 0.0 | 2.2 | 0.9 | 0.0 | 3.5 | 2.5 | 0.9 |
| 3 | 5.3 | 2.6 | 1.6 | 5.4 | 3.2 | 2.0 | 5.5 | 4.7 | 4.1 |
| 4 | 8.7 | 7.1 | 5.2 | 7.8 | 7.3 | 6.3 | 8.6 | 8.1 | 8.2 |
| 5 | 9.7 | 9.5 | 8.8 | 9.5 | 9.5 | 8.7 | 9.7 | 9.6 | 9.5 |

In Table 1, the results from Hart and Ross managed to create a trend where the average number of antigens not matched by any antibody decreases as the size of the antibody samples, $s$ increases from 5 to 30. The results in Table 2 are in line with the trend where the average number of unmatched antigens still decreases when the whole population is matched against the antigens. However, the main difference between the results compared to Hart and Ross's was that as we increase the number of antigens, the average number of antigens that cannot be matched by any antibody decreases. While the result in [17] could be interpreted as evidence that more specific antibodies have been produced, we believe that, as we expose more antigens to the antibodies, the fitness of the antibodies would increase and therefore would result in more antigens getting matched or recognized. Therefore with our model, we can produce partial schedules that can be used as replacement to an actual schedule when disturbances occur.

**Table 2.** Average number of antigens (out of a possible 10) not matched by any antibody (modified algorithm for AIS)

| Match Thres-hold | Ab = 100 | | | | | | | | |
|---|---|---|---|---|---|---|---|---|---|
| | Type A | | | Type B | | | Type C | | |
| | Ag | | | Ag | | | Ag | | |
| | 1 | 4 | 8 | 1 | 4 | 8 | 1 | 4 | 8 |
| 2 | 0.0 | 0.0 | 0.0 | 0.0 | 0.0 | 0.0 | 0.0 | 0.0 | 0.0 |
| 3 | 0.4 | 0.0 | 0.0 | 0.9 | 0.1 | 0.0 | 0.8 | 0.1 | 0.0 |
| 4 | 6.5 | 3.6 | 1.3 | 6.2 | 3.4 | 1.4 | 6.6 | 3.2 | 1.3 |
| 5 | 8.5 | 6.3 | 4.7 | 8.3 | 6.6 | 5.3 | 8.2 | 7.1 | 5.8 |

We also ran experiments to see if our Phase II could improve the results. Two different sets of experiments have been carried out, where we use the final populations generated using our artificial immune system as initial solutions to the simulated annealing and the great deluge algorithms separately (Phase II). Ten different sets of antibody populations (initial solutions) were used for each sample of antigens. The final populations generated from this phase were then matched with all the existing ten antigens to illustrate the diversity of the antibodies/partial schedules created. Two different operators were tested. As the operator swapping two jobs generated similar results, we present only the results tested with the operator of changing one job in each antibody.

The results depicted in Table 3 are the average number of antigens not matched by any antibody for both hybrid models compared to the artificial immune system alone with a mutation rate of 0.001 on antibody population Type A. We also show the percentage of the fitness improvement on antibodies generated using the hybrid search algorithm compared with the fitness of the antibodies generated using our new artificial immune system algorithm in the table. It is important to note that the time taken to generate initial antibody populations is less than one minute. The time taken to get a final population (antibody population) using our artificial immune system algorithm (from Phase I) is between one to two minutes while the time taken to get a final solution (antibody population) using a hybrid with the simulated annealing and great deluge algorithms, respectively (Phase II) is one minute or less. This applies to any parameter used to evolve the final populations. We believe this is due to the cooling schedule that is used in the simulated annealing algorithm and the number of iterations set in the great deluge algorithm.

**Table 3.** Average number of antigens (out of a possible 10) not matched by any antibody in population Type A (for New AIS (our artificial immune system), AIS+SA (our artificial immune system hybridised with the simulated annealing algorithm) and AIS+GD (our artificial immune system hybridised with the great deluge algorithm))

| Match Thres-hold | Ab = 100 | | | | | | | | |
|---|---|---|---|---|---|---|---|---|---|
| | **New AIS** | | | **AIS + SA** | | | **AIS + GD** | | |
| | Ag | | | Ag | | | Ag | | |
| | 1 | 4 | 8 | 1 | 4 | 8 | 1 | 4 | 8 |
| **2** | 0.0 | 0.0 | 0.0 | 0.0 | 0.0 | 0.0 | 0.0 | 0.0 | 0.0 |
| **3** | 0.6 | 0.1 | 0.0 | 1.5 | 0.3 | 0.0 | 1.4 | 0.4 | 0.0 |
| **4** | 6.8 | 3.0 | 1.0 | **6.6** | 4.7 | **0.6** | 6.5 | 4.0 | 1.4 |
| **5** | 7.9 | 6.0 | 4.4 | 8.3 | **5.3** | 3.5 | 8.2 | **5.1** | 4.8 |
| Fitness Diff. (%) | | | | 28.5 | 11.7 | 4.8 | 28.8 | 10.7 | 4.7 |

The results of the experiments show, not surprisingly, that the hybrid search algorithms do improve on the artificial immune system algorithm developed in [17] and in this research. However, the hybrid algorithm does not improve the coverage of the antigen universe compared to our artificial immune system algorithm alone except for certain combinations of the number of antigens and the matching threshold. This is probably due to the large number of general antibodies (partial schedules) produced using the artificial immune system that can be matched with most of the antigens. Both hybrid models produced more specific antibodies and, therefore, could not cover most of the antigens.

The fitness of the antibodies in the population, however, does improve, as depicted in the last row in Table 3. Here we total up the fitness of all the antibodies in all ten different sets of antibody populations for each sample of antigens for both hybrid search algorithms and our artificial immune system algorithm. The fitness of the whole antibody populations generated using the hybrid simulated annealing and the hybrid great deluge algorithm, respectively increases by more than 28% over the antibodies using the artificial immune system alone. However, the percentage drops gradually as the number of antigens selected increases.

## 5 Conclusion

This paper has solved a simple job shop scheduling problem. We have developed an artificial immune system model by drawing upon the research in [17,18]. Our empirical results represent an improvement upon those in [17,18]. We also investigated the use of local search methods to further improve the partial schedules developed in the antibody population. The results obtained indicated that the hybridisation of our artificial immune system approach with simulated annealing and great deluge, respectively, did not yield improvement in terms of the coverage of the antigen universe. However, they did improve the fitness of the antibodies produced in the population. This is important, as we need to provide a range of good partial schedules that can be used to replace certain jobs in the actual schedule when we have changes in the arrival dates of the jobs. We will also use the results as a platform for our future work on hyper heuristic. For the problem, the antibodies will represent a sequence of low level heuristic.